\documentclass[letterpaper, 10 pt, conference]{ieeeconf}  

\IEEEoverridecommandlockouts                              

\usepackage{graphics} 
\usepackage{epsfig} 
\usepackage{mathptmx} 
\usepackage{times} 
\usepackage{amsmath} 
\usepackage{amssymb}  
\usepackage{multirow}
\usepackage{breqn}
\usepackage{xcolor}

\title{\LARGE \bf
Relevance-driven Decision Making for Safer and More Efficient Human Robot Collaboration
}

\author{Xiaotong Zhang, Dingcheng Huang and Kamal Youcef-Toumi
\thanks{All authors are with the Mechatronics Research Laboratory, Massachusetts Institute of Technology, Cambridge, MA, 02139, USA
{\tt\small \{kevxt, dean1231, youcef\}@mit.edu }}
\thanks{This research was made possible by the support and partnership
of King Abudlaziz City for Science and Technology
(KACST) through the Center for Complex Engineering
Systems at Massachusetts Institute of Technology (MIT) and
KACST.}
}

\begin{document}

\maketitle
\thispagestyle{empty}
\pagestyle{empty}

\begin{abstract}
Human brain possesses the ability to effectively focus on important environmental components, which enhances perception, learning, reasoning, and decision-making. Inspired by this cognitive mechanism, we introduced a novel concept termed relevance for Human-Robot Collaboration (HRC). Relevance is a dimensionality reduction process that incorporates a continuously operating perception module, evaluates cue sufficiency within the scene, and applies a flexible
formulation and computation framework. In this paper, we present an enhanced two-loop framework that integrates real-time and asynchronous processing to quantify relevance and leverage it for safer and more efficient
human-robot collaboration (HRC). The two-loop framework integrates an asynchronous loop, which leverages an LLM’s world knowledge to quantify relevance, and a real-time loop, which performs scene understanding,
human intent prediction, and decision-making based on relevance. HRC decision-making is enhanced by a relevance-based task allocation method, as well as a motion generation and collision avoidance approach
that incorporates human trajectory prediction. Simulations and experiments show that our methodology for relevance quantification can accurately and robustly predict the human objective and relevance, with an average accuracy of up to 0.90 for objective prediction and up to 0.96 for relevance prediction. Moreover, our motion generation methodology reduces collision cases by 63.76\% and collision frames by 44.74\% when compared with a state-of-the-art (SOTA) collision avoidance method. Our framework and methodologies, with relevance, guide the robot on how to best assist humans and generate safer and more efficient actions for HRC. 
\end{abstract}


\section{Introduction}
Robotic and automation systems are becoming increasingly critical in enhancing productivity, efficiency, and precision in industry \cite{zhang2022magnetohydrodynamic, zhang2022systematic,zhang2019design} and daily human life \cite{kothari2023enhanced,xia2021modular}. However, they still struggle to match the exceptional cognitive capabilities of human beings. Among all cognitive mechanisms in human brains, humans are extraordinary at selectively focusing on relevant stimuli while filtering out irrelevant information in our environment, which plays a crucial role in more efficient spatial perception, scene understanding, and decision-making \cite{bundesen1990theory, yoo2022brain}. This mechanism is regulated by a structure of neurons in the brainstem, known as the Reticular Activating System (RAS), which filters sensory information and determines its relevance. The earliest paper on RAS in the human brain can be
traced to the work in \cite{moruzzi1949brain}. It laid the foundation for
understanding the RAS and its role in consciousness,
attention, and alertness.

To empower robots with similar capabilities, in \cite{zhang2024relevancehumanrobotcollaboration}, we defined a new and important concept and dimensionality reduction approach, termed relevance, and developed methodologies for quantifying relevance with a novel event-based framework and a probabilistic methodology based on a new scene representation. Relevance is
defined as a
dimensionality
reduction process
that continuously
perceives the
scene, identifies
and detects its
elements, and
organizes them
into a hierarchical
representation,
such as attribute
classes. This
process utilizes
contextual cues
to selectively
reduce input
dimensionality.
When cues are
insufficient,
relevance-driven
processing
iteratively gathers
additional cues
and reapplies
dimensionality
reduction until
sufficient
information is
obtained. 

Relevance offers three key benefits in human-robot collaboration (HRC). First, accurately determined relevance enables the robot to better understand object applicability to the scene and potential interaction sequences, enhancing the efficiency, safety, and fluency of HRC. Second, focusing on relevant objects allows the robot to optimize its computational resources, improving speed and safety \cite{zhang2024does}. Third, relevance unites human factors, task models, scene relationships, and even information sufficiency, facilitating more accurate predictions and reasoning by leveraging advancements from various areas.


Building on the work in \cite{zhang2024relevancehumanrobotcollaboration}, this paper further develops and presents a framework to leverage an AI toolkit to determine relevance and apply it to a novel real-time decision-making component that enhances safety in HRC. In this framework, two loops are running asynchronously, namely the async loop and the real-time loop. The real-time loop performs continuous real-time computation, including scene understanding, human intent prediction, and decision-making. The asynchronous loop leverages the world knowledge from Large Language Models (LLM) in the AI toolkit to predict the potential human objective and relevance. The real-time loop utilizes those results for decision-making to achieve proactive and safe HRC. The execution rates of the two asynchronous loops adjust dynamically based on the scene's conditions.

In the decision-making module, we developed a real-time methodology with human-robot task allocation and a novel motion planning methodology based on relevance. The traditional Artificial Potential Field (APF) methods for motion planning and collision avoidance are enhanced by a virtual obstacle constructed based on the future projection of the human motion and associated repulsive force formulation. In this way, the robot can behave in a proactive manner for safer and more efficient HRC. Simulation results demonstrate that our decision making module reduces the collision cases and frames by 63.76\% and 44.74 \%, respectively. 


In summary, the paper's contributions are as follows: (1) We proposed and developed a novel methodology to quantify relevance with human objective prediction and relevant object prediction based on an AI toolkit. (2) We proposed and developed a novel two-loop framework to integrate the LLM inference in the AI toolkit and relevance quantification into a real-time application by leveraging the asynchronous frameworks based on different inference requirements of different components. (3) We proposed and developed a novel methodology of decision-making based on relevance. Simulations and experiments validate enhanced safety and lower collision probability with our Relevance-based Artificial Potential Field (RAPF).


\section{Related Works}
This work is unique to the best of our knowledge and exceptionally contributes to the area of robotics in the following manners.

\subsection{Saliency and Attention}
Relevance is a broad concept that covers previous works identifying and highlighting the important features in the scene, such as attention and saliency. Saliency focuses on identifying the visually prominent and conspicuous features in the image \cite{itti1998model,meger2008curious,chung2002new}, while attention focuses on current short-term, reactive, and simple tasks without considering the future projection \cite{vaswani2017attention,devin2018deep}. Relevance possesses more advanced functionalities by considering information sufficiency, hierarchical scene representation, dynamically scheduled perception modules, etc. Furthermore, relevance is characterized by a more flexible formulation and computational principle than existing methods, which results in unique and additional advantages leading robots toward Artificial General Intelligence (AGI). A more detailed comparison can be found in \cite{zhang2024relevancehumanrobotcollaboration}.

\subsection{LLMs in the AI Toolkit for Robotics}
Large Language Models (LLMs) are revolutionizing robotics by enabling multi-modality reasoning \cite{yang2024embodied}, general prediction without transfer learning \cite{chen2024driving}, and flexible connections between modules in robotic frameworks \cite{zhou2024isr}. In this paper, we utilized LLMs in the AI toolkit for a dramatically different and novel application, i.e., relevance determination. Moreover, the current inference time of LLMs is longer than that of real-time computation requirements. Existing works with LLMs in robotics mainly apply to non-real-time policy generation in a static or pre-defined environment. Our unique two-loop design enables knowledge retrieval of LLMs in real-time applications by leveraging asynchronous computation. 

\subsection{Artificial Potential Fields}
APF is an intuitive and efficient real-time robot motion planning method by simulating attractive forces toward goals and repulsive forces away from obstacles, enabling smooth and collision-free navigation \cite{khatib1986real}. Song et al. proposed a methodology called Predictive APF to anticipate obstacles based on the robot's velocity and the relative positions of obstacles to adjust the path before potential collisions \cite{song2020path}. However, their goal of incorporating prediction into APF is to smooth the path considering the dynamic constraints of the agent, which is dramatically different from ours. Moreover, their methodology is limited to environments with static obstacles. Our Relevance-based APF (RAPF) is to improve HRC safety by predicting the motion of the human and dynamically updating the path proactively in a highly dynamic environment.

\section{Problem Definition and Methodology}
In this section, we introduce the problem definition and methodology for quantifying relevance and applying it to sensorimotor policy generation for safer and more efficient HRC.

\subsection{Problem Definition}
Let $S = \{C_{1}, C_{2}, ..., C_{n}\}$ be a set of class of objects in a visual scene, where each $C_{i}$ represents a class of objects and $n$ is the number of classes of objects in the scene. 
The set of relevant classes $\mathcal{C}_r$ and the set of relevant elements $\mathcal{E}_r$, based on the human's objective and associated tasks, is predicted and determined with the available information. Based on $\mathcal{C}_r$ and $\mathcal{E}_r$, the robot generates actions to assist the human in achieving the human's objective.  



\subsection{Framework Overview}

\begin{figure*}[!t]
\begin{center}
\includegraphics[width=0.9\textwidth]{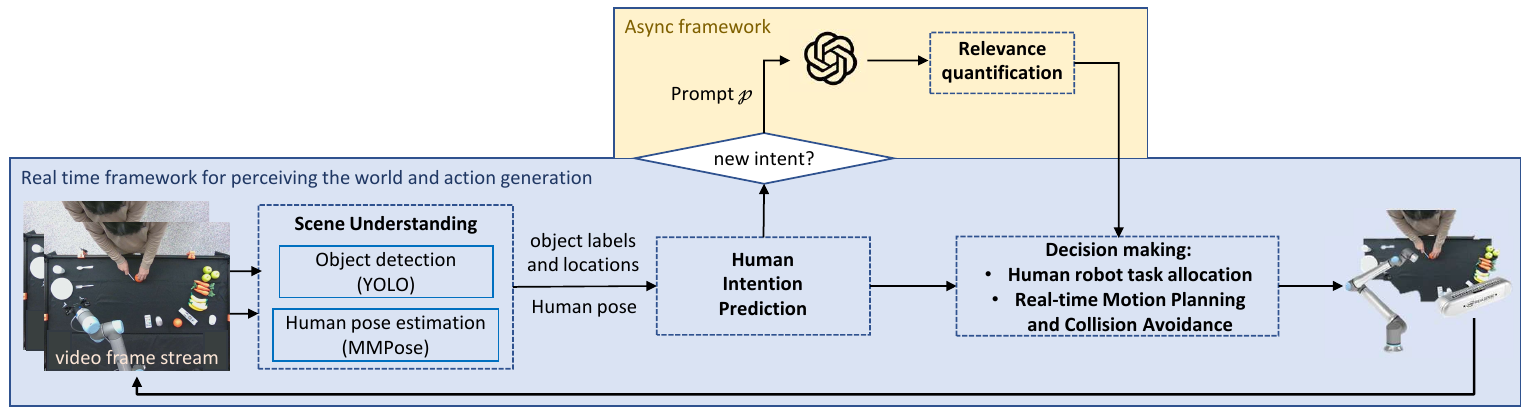}
\end{center}
\caption{The overview of the methodology for relevance quantification and application of relevance for safer human-robot collaboration. The inputs to the algorithm pipeline are the stream images from the camera, which are processed with scene understanding algorithms in the AI toolkit. Based on the scene understanding, the human intent is classified and fed into the LLM within the AI toolkit in an asynchronous manner. With the LLM results, the relevance can be quantified, which informs the decision-making module to generate actions for the robot for safer HRC.}
\label{fig: methodology}
\end{figure*}

An overview of the framework is shown in Fig. \ref{fig: methodology}. Our framework for relevance quantification and safer HRC consists of two loops. The first loop fulfills the function of perceiving the environment, understanding the scene, and decision making based on relevance for safer HRC. The second loop is an asynchronous loop that leverages modules in an AI toolkit and quantifies relevance. In this paper, we leverage an LLM for the purpose of illustration. By optimally coordinating the two loops, the world knowledge of the LLM in the AI toolkit can be incorporated into real-time operation of the robots. 

\subsection{Scene Understanding}
The input to the algorithm is the stream of observations from the robot sensors. The images can be processed with scene understanding algorithms in the AI toolkit, such as YOLO \cite{jiang2022review} for object detection, MMPose \cite{mmpose2020} for human pose estimation, CLIP \cite{radford2021learning} for scene grounding, etc. 

\subsection{Human Intention Prediction}
The module of human intention prediction determines current human actions and aggregates them into a history record. Here, we implement an algorithm that leverages Bayesian networks to fuse three modalities of human information: head orientation, hand orientation, and hand movement \cite{hernandezcruz2024bayesianintentionenhancedhuman}. By fusing three modalities for human intention recognition, the accuracy and precision of the prediction are dramatically increased because different modality provides valuable information in different stages of the actions. 

\subsection{Relevance Quantification with LLMs}
In the original relevance formulation and quantification methodology, the scene is represented as classes and elements. The relevance of each class and element is determined in a sequential and hierarchical manner. In this paper, we adopted a direct prediction based on LLMs on the element level for simplicity. The goal of the LLMs in this paper is to predict the human objective and the set of relevant elements $\mathcal{E}_r$, reflecting potential future object interactions and human motion.

The first step in quantifying relevance with LLM is contextualizing the scene. Currently, this is achieved with a fixed prompt $\mathcal{P}$ designed and optimized specifically for the setup and the problem of HRC. However, in the future, automatic scene contextualization and prompt construction can be considered and developed. 
In the prompt $\mathcal{P}$, the environment \textit{env}, the objects in the scene \textit{object in env}, and the action history of the human \textit{action history} are automatically generated from the scene understanding and human intention prediction modules. An example of the prompt is: \textit{In a kitchen with [object in env], a person has already grabbed [action history]. What objective is the human trying to finish? What objects will be relevant next among [object in env]?} In the prompt, \textit{[object in env]} and \textit{[action history]} can be derived and populated automatically. The output format regulation of the prompt is omitted due to the page limit. 


The outputs of the LLM module are the prediction of the human objective and the set of relevant elements $\mathcal{E}_r$. With this method, world knowledge of an LLM can be extracted for a general environment and setup without any training and transfer learning.

\subsection{Decision Making - Task Allocation}
Our framework's decision-making can be decomposed into two components: human task allocation and real-time motion generation with collision avoidance. 

For the sake of simplicity, we assume that humans need all the relevant elements predicted for the objective. A more detailed inquiry method to derive the necessary and preferred elements can be found in \cite{zhang2024relevancehumanrobotcollaboration}. The human-robot task allocation is formulated as an optimization problem and solved with an optimization solver. 


In the problem definition, we define the subscript $j$ representing the index of elements such that $e_j\in \mathcal{E}_r$. We define the position vector of element $e_j$ in the world coordinate as $\mathbf{p}_{e_j}$, the position of the common destination of relevant elements as $\mathbf{p}_d$, the initial location of the robot as $\mathbf{p}_{r_0}$, the velocity of the robot as $v_r$, the velocity of human as $v_h$, the start-up delay for the human to finish the current task as $d_h$. We consider $d_h$ because there is a temporal delay between the time the robot and the human start to fulfill the allocated tasks. We further define $T_r$ and $T_h$ as the time durations for the robot or the human to finish the assigned tasks, which can be computed as: 

\begin{equation}
\begin{aligned}
T_r &= \sum_{j} y_j \left(\frac{\|\mathbf{p}_{r_0} - \mathbf{p}_{e_j}\| + \|\mathbf{p}_{e_j} - \mathbf{p}_d\|}{v_r}\right) \\
& \quad + \sum_{j} x_j (1 - y_j) \left(\frac{2\|\mathbf{p}_d - \mathbf{p}_{e_j}\|}{v_r}\right)
\end{aligned}
\end{equation}

and

\begin{equation}
T_h = d_h + \sum_{j} (1 - x_j) \left(\frac{2\|\mathbf{p}_d - \mathbf{p}_{e_j}\|}{v_h}\right)
\end{equation}

We define the decision variable $Z$ as the maximum time taken to complete all tasks, and the optimization problem is formulated as:

\begin{IEEEeqnarray}{lll}
\text{Minimize} & \quad & Z \IEEEyesnumber \label{eq:minimize} \\
\text{Subject to:} \nonumber \\
& \sum_{j} y_j &= 1, \IEEEyesnumber \label{eq:sum_yi} \\
& y_j &\leq x_j \quad \forall j, \IEEEyesnumber \label{eq:yi_xi} \\
& Z &\geq T_r, \IEEEyesnumber \label{eq:Z_Tr} \\
& Z &\geq T_h, \IEEEyesnumber \label{eq:Z_Th} \\
& x_j, y_j &\in \{0, 1\} \quad \forall j. \IEEEyesnumber \label{eq:binary}
\end{IEEEeqnarray}



where $x_j$ is a binary decision variable where $x_j = 1$ if the robot moves $e_j$, and $0$ otherwise, and $y_j$ is a binary variable where $y_j = 1$ if $e_j$ is the first task done by the robot, and $0$ otherwise. (\ref{eq:sum_yi}) ensures that exactly one object is designated as the initial task performed by the robot. (\ref{eq:yi_xi}) specifies that an object can only be the first task executed by the robot if that object is assigned to the robot, as indicated by $x_j$. After the relevant objects are allocated to the robot or the human, the robot will execute the assigned tasks, generate the motion, and avoid obstacles dynamically in real time. 

\subsection{Decision Making - Motion Planning}

\begin{figure}[!t]
\begin{center}
\includegraphics[width=0.8\columnwidth]{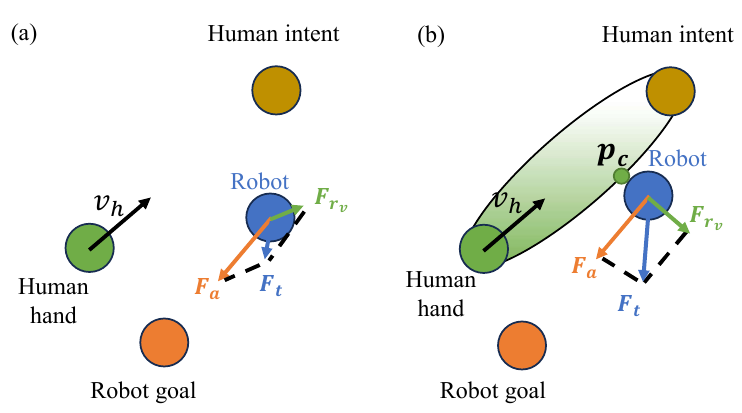}
\end{center}
\caption{Illustration of (a) normal APF; (b) virtual obstacle and the updated repulsive force. By constructing a virtual ellipsoid obstacle from the current location of the hand to the destination of the human hand, the repulsive force is more proactive, which results in safer collision avoidance.}
\label{fig: collision avoidance}
\end{figure}

For dynamic motion planning and obstacle avoidance, our methodology is based on an APF formulation in \cite{8460185}. The major contribution of our work, as shown in Fig. \ref{fig: collision avoidance}, is that we build a virtual obstacle from the current human hand based on the predicted human motion and update the repulsive force based on the virtual obstacle for proactive and safer motion generation.

In our APF, the attractive force $\mathbf{F}_a$ is modeled as: 

\begin{equation}
\mathbf{F}_a = A \left( 1 - \exp\left(-\frac{d_g} {\alpha_a}\right) \right) \mathbf{\hat{u}_a}
\end{equation}

where $A$ is the magnitude of the attractive force, $d_g$ is the distance between the robot and its goal, $\alpha_a$ is the constant that controls how the attractive force increases with distance, and $\mathbf{\hat{u}_a}$ is the unit vector from the robot to the goal. 

The repulsive force by obstacles other than the human's hand $\mathbf{F}_{r_o}$ is constructed as: 

\begin{equation}
\mathbf{F}_{r_o} = \sum_{i=1}^m \frac{R}{1 + \exp\left( \left( \frac{2d_{e_i}}{\rho_r} - 1 \right) \alpha_r \right)} \mathbf{\hat{u}_{r_i}}
\label{eq: rep force}
\end{equation}

where $m$ is the number of elements in the scene, $R$ is the magnitude of the repulsive force, $d_{e_i}$ is the distance between the robot and the obstacle $e_i$, and $\mathbf{\hat{u}_{r_i}}$ is the unit vector from the obstacle $e_i$ to the robot location. $\rho_r$ and $\alpha_r$ are two factors defining the shape and decreasing rate of the repulsive force as the distance increases, respectively.

To consider the future motion of the human hand, we developed a novel methodology to construct an ellipsoid virtual obstacle, as shown in Fig. \ref{fig: collision avoidance}. The two endpoints of the major axis of the ellipsoid are the current position of the human hand \( \mathbf{p}_h \) and the position of the predicted hand destination \( \mathbf{p}_t \). The semi-axes \( a \), \( b \), and \( c \) of the ellipsoid are given by:


\begin{equation}
a = \frac{\|\mathbf{p}_h - \mathbf{p}_t\|}{2}
\end{equation}
\begin{equation}
b = k_b \cdot a
\end{equation}
\begin{equation}
c = k_c \cdot a
\end{equation}

where $k_b$ and $k_c$ are two factors controlling the shape of the ellipsoid. The repulsive force from this virtual obstacle is updated to

\begin{equation}
\mathbf{F}_{r_v} = \frac{k_r \cdot R}{1 + \exp\left( \left( \frac{2d_{c}}{\rho_r} - 1 \right) \alpha_r \right)} \mathbf{\hat{u}_{r_v}}
\end{equation}

Where $d_c$ is the distance from the robot to the closest point on the ellipsoid $\mathbf{p}_c$, $\mathbf{\hat{u}_{r_v}}$ represents the unit normal vector at the closest point on the ellipsoid, and $k_r$ is a new factor we proposed representing a scale factor on the force magnitude based on the proximity to the human. In this paper, $k_r$ is computed as: 

\begin{equation}
k_r = 1 - \frac{(\mathbf{p}_c - \mathbf{p}_h)\cdot \mathbf{\hat{u}_{h\rightarrow t}}}{(v_h+v_r)\cdot t_s}
\end{equation}

where $\mathbf{\hat{u}_{h\rightarrow t}}$ represents the unit vector from $\mathbf{p}_h$ to $\mathbf{p}_t$, and $t_s$ is a time factor we proposed in the unit of $s$ reflecting the available safety buffer in terms of time. $k_r$ decreases as $\mathbf{p}_c$ becomes farther away from $\mathbf{p}_h$. Without $k_r$, $\mathbf{F}_{r_v}$ will consider the virtual obstacle too early and conservatively, resulting in unnecessary detours and a longer robot trajectory.


The total force acting on the robot $\mathbf{F}_t$ is the sum of the attractive force and repulsive forces from both the physical obstacles and the virtual obstacle:

\begin{equation}
\mathbf{F}_{t} = \mathbf{F}_a  + \mathbf{F_{r_o}} + \mathbf{F}_{r_v}
\end{equation}

This total force determines the direction and magnitude of the robot's motion, allowing it to avoid both physical and virtual obstacles while moving toward its goal.

\section{Evaluation Setup}
In this section, we introduce the evaluation setup for our proposed methodologies on leveraging LLMs for objective and relevance prediction and relevance in decision making for safer and more efficient HRC. 

\subsection{Dataset}
The performance of human objective prediction and relevance prediction is assessed via the Breakfast Actions Dataset \cite{Kuehne12}, which comprises a variety of typical human activities performed during breakfast time (e.g., preparing coffee, cooking pancakes, making hot chocolate, etc.). For each test, ground truth data provides the ground truth objective (GTO), and the ground truth plan (GTP), which describes the sequence of actions performed by an individual in the execution of the objective. 
In our evaluation, a step ratio of segments from the GTP is contextualized and fed into LLM to predict the human objective and the set of relevant elements. Three distinct ratio step values, 0.25, 0.5, and 0.75, are employed. The objective prediction is evaluated through a manual assessment process. The predicted relevant objects are evaluated automatically with word matching. The metrics for the evaluation are percentages of tests with correct prediction of the objective and the relevant elements.  


\subsection{Simulation Development and Setup}
\begin{figure}[!h]
\begin{center}
\includegraphics[width=0.6\columnwidth]{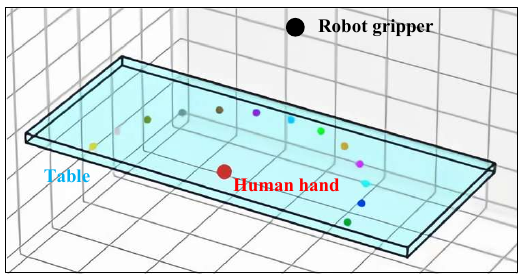}
\end{center}
\caption{The simulation setup for a test case of \textit{making cereals}. The human's hand (red dot) starts to move to the human's first intent. The UR5 robot, whose gripper is shown with the black dot, observes the human's actions, quantifies relevance, and makes informed decision-making for safer collision avoidance. 13 objects for the problem are randomly and uniformly distributed along a half circle on the table. }
\label{fig: simulation_setup}
\end{figure}

To verify the effectiveness of relevance for safer HRC, we developed a simulator as shown in Fig. \ref{fig: simulation_setup}. A large number of simulations can be carried out systematically and identically for different test cases, facilitating the evaluation of the decision making module. In the simulator, we constructed the environment with a table (the cyan cuboid) and objects on top of the table. The height of the table is set to be 73 cm. The table size is 180 cm in the $x$ direction, 6 cm in the $z$ direction, and 76 cm in the $y$ direction. The objects on the table consist of two parts: the necessary objects for the objective and randomly added objects from a list of kitchen objects. The locations for each object on the tabletop are randomly selected from a collection of locations uniformly distributed on a half circle with a radius of 60 cm. This object arrangement enables a wide range of relative velocities and relative angles between moving objects during testing, recovering realistic scenarios. The diameter of each object is 8 cm. 

A human's hand (the red dot) is added to the simulator to simulate the human's motion. A UR5 robot is mounted on the other end of the table, reasoning about the human objective and the relevant objects and generating safe actions to best assist the human by accomplishing the human's objective with minimum time. The initial locations of the human and the robot are shown in Fig. \ref{fig: simulation_setup}. To take the inference time of LLM into consideration, the simulation runs and progresses in real time with a frequency of 30 Hz. The velocities of the human's hand and the end gripper are set to be 0.4 $m/s$. 

To test the effectiveness of relevance and our decision-making methodology, we focus on comparing two methods. The first method uses relevance and our decision-making with virtual obstacles, as described in Section III. For the comparison, we built a baseline test without virtual obstacles but with relevance. It's worth noticing that, without relevance, the robot has no idea about its tasks to assist the human. Thus, we assign the same tasks as the relevance test cases to the robot. And the robot starts to move with exactly the same frame index as the relevance cases. Without loss of generality and for simplicity, we test on the objective of \texttt{making cereals}, and the relevant objects are \texttt{cereal, bowl, milk, and spoon}. 

\section{Evaluation Results}
In this section, we present the evaluation results of our proposed methodologies, demonstrating their effectiveness. 

\subsection{Accurate Objective and Relevance Prediction}

Our evaluation results are depicted in Table. \ref{table: llm results}, confirming the effectiveness of our methodology for objective and relevance prediction across various scenarios. At a step ratio of 0.75 with more human actions contextualized and fed into LLM, our methodology achieves a high objective prediction accuracy of 0.90 and a relevance prediction accuracy of 0.96.

\begin{table*}[t]
\caption{Performance of objective prediction and relevance prediction for the 10 objectives in the Breakfast Actions Dataset with various step ratios of action as inputs. These results show that our methodology can accurately and robustly predict human objectives and relevance.}
\centering
\begin{tabular}{ccccccccccccc}
\hline
\multirow{2}{*}{}                                                                & \multirow{2}{*}{\begin{tabular}[c]{@{}c@{}}Step \\ Ratio\end{tabular}} & \multirow{2}{*}{Average} & \multicolumn{10}{c}{Objectives}                                                                         \\ \cline{4-13} 
                                                                                 &                                                                        &                          & Cereals & Coffee & Friedegg & Chocolate & Juice & Pancake & Salad & Sandwich & Scrambledegg & Tea  \\ \hline
\multirow{3}{*}{\begin{tabular}[c]{@{}c@{}}Objective \\ prediction\end{tabular}}      & 0.25                                                                   & 0.69                     & 0.93    & 0.96   & 0.16     & 0.76      & 0.72  & 0.43    & 0.92  & 0.65     & 0.42         & 0.92 \\
                                                                                 & 0.5                                                                    & 0.82                     & 1.00    & 1.00   & 0.13     & 1.00      & 0.96  & 0.95    & 1.00  & 0.32     & 0.84         & 0.96 \\
                                                                                 & 0.75                                                                   & 0.90                     & 1.00    & 1.00   & 0.28     & 1.00      & 0.92  & 1.00    & 1.00  & 0.88     & 0.95         & 1.00 \\ \hline
\multirow{3}{*}{\begin{tabular}[c]{@{}c@{}}Relevance \\ Prediction\end{tabular}} & 0.25                                                                   & 0.77                     & 0.78    & 0.56   & 0.66     & 1.00      & 1.00  & 0.67    & 1.00  & 0.88     & 0.68         & 0.50 \\
                                                                                 & 0.5                                                                    & 0.94                     & 0.96    & 0.96   & 0.88     & 1.00      & 1.00  & 0.81    & 1.00  & 1.00     & 1.00         & 0.75 \\
                                                                                 & 0.75                                                                   & 0.96                     & 1.00    & 0.88   & 0.88     & 1.00      & 1.00  & 0.95    & 1.00  & 0.97     & 0.95         & 0.96 \\ \hline
\end{tabular}
\label{table: llm results}
\end{table*}

We first analyze the performance of objective prediction using our methodology, upon which relevance determination depends. Upon further examination of individual objectives, certain objectives, such as the preparation of cereal, coffee, hot chocolate, juice, salad, and tea, exhibit consistently high predictive accuracy across all evaluated ratio steps. The prediction accuracy for those objectives is high even at a low step ratio of 0.25, demonstrating our methodology can accurately identify the human's objective at an early stage of the human action and motion. Consistent with expectations, an increase in the ``ratio steps" parameter correlates positively with enhanced prediction accuracy. The contextualization of the scene with a higher step ratio will contain more indicative cues about the objectives and thus improve the objective prediction accuracy. 
The inaccuracy for fried eggs is attributed to the inherent similarity between the objectives involving eggs, such as omelettes and scrambled eggs. However, the relevance for those objectives is very similar. 

Next, we analyze the performance of relevance prediction using our methodology based on the predicted objective. The relevance prediction performs extraordinarily for specific objectives, including hot chocolate, juice, and salad, which achieves a 100\% accuracy at the step ratio of 0.25. The prediction accuracy of all other objectives achieves at least 0.85 at the step ratio of 0.75. For the relevance prediction, the accuracy also increases with the step ratio. 
It's worth noting that the relevance prediction for the objectives predicted with low accuracy can still be accurate. This is attributed to the fact that
similar objectives, such as omelettes and scrambled eggs, share the same relevance, which bolsters the relevance prediction despite inaccuracies in objective identification.

\subsection{Real-time and Safer Decision Making}
Through all the test cases, our methodology of relevance determination robustly and correctly predicts the human's objective and the set of relevant elements for making cereals to be bowl, spoon, milk, and cereal. Our human robot task allocation module robustly solves the optimization problem in real-time to generate the tasks for the robot. Thus, we focus on the evaluation of the virtual obstacle and collision avoidance. The test results of our real-time and safe decision making methodology are shown in Table \ref{table:collision_statistics}. 

\begin{figure}[!h]
\begin{center}
\includegraphics[width=\columnwidth]{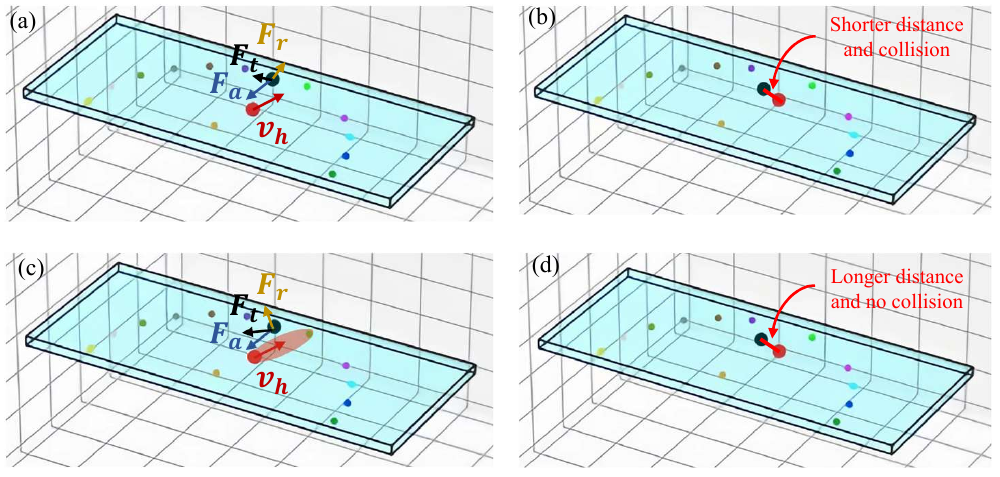}
\end{center}
\caption{Visualization of motion generation of (a) baseline at $t=3.37s$; (b) baseline at $t=3.67s$; (c) our methodology at $t= 3.37s$; (d) our methodology at $t=3.67s$. The gold arrow, blue arrow, black arrow, and red arrow represent the repulsive force generated by the hand, the attractive force, the total force, and the hand velocity, respectively. The sizes of the dots are not scaled to the real size in the simulator. Thus, two red lines with the same length are shown in (b) and (d) for the detailed comparison of the distances between two objects. By introducing the virtual obstacle and the associated repulsive force formulation, the repulsive force generated by the human hand is more perpendicular to the anticipated human hand trajectory, pushing the gripper farther away from the human hand.}
\label{fig: Collision_visualization}
\end{figure}

With our RAPF, the robot actions generated are much safer, with the rate of collided cases decreasing by 63.76\% and the rate of collided frames decreasing by 44.74\%. A visual comparison between the two methods is shown in Fig. \ref{fig: Collision_visualization}. The frames at $t=3.37s$ are the frames before the collision. With the relevance and the virtual obstacle, the repulsive force (shown with a gold arrow) is more perpendicular to the hand trajectory and thus pushes the gripper further from the hand smoothly. At $t=3.67s$, the gripper and the human collide in the baseline test because there is not enough duration of repulsive force to push the gripper farther away from the hand trajectory. However, with the virtual obstacle, no collision happens because of the proactive repulsive force generated by the virtual obstacle. Those results demonstrate the effectiveness of relevance and RAPF in informing the robot how to best assist humans and generate safer and more efficient actions.

\begin{table}[]
\caption{Comparison of Collision Rate. With relevance and RAPF, the safety of HRC can be dramatically improved.}
\centering
\begin{tabular}{cccc}
\hline
                        & Baseline & RAPF           &  Percentage decrease\\ \hline
Rate of collided cases  & 0.149    & \textbf{0.054} & 63.76\%             \\
Rate of collided frames & 0.010    & \textbf{0.006} & 44.74\%             \\ \hline
\end{tabular}
\label{table:collision_statistics}
\end{table}


\section{Conclusions}
Relevance is a novel concept inspired by human cognitive abilities,
enabling adaptive dimensionality reduction of a scene based on objectives,
context, or other influencing factors. In this paper, we developed a novel two-loop framework that robustly, efficiently, and accurately quantifies relevance and applies relevance to improve the effectiveness and safety of HRC. This framework integrates an asynchronous loop to leverage world knowledge from an AI toolkit and quantify relevance, as well as a real-time loop to execute scene understanding, human intention prediction, and decision-making. Moreover, we proposed and developed a decision making methodology based on relevance, integrating human robot task allocation and real-time motion generation. In motion generation, we developed a methodology to construct a virtual obstacle and formulate the associated repulsive force. Simulations and experiments verify that our framework performs well for relevance quantification, with an objective prediction accuracy of 0.90 and a relevance prediction accuracy of 0.96. Simulations further verify our novel motion generation methodology dramatically decreases the cases with a collision by 63.76\% and the frames with a collision by 44.74\%. The robot is comprehensively informed about how to best assist the human with relevance, and our decision making module generates safe actions to achieve the assistance, leading robotics toward artificial general intelligence.


\addtolength{\textheight}{-8cm}   




\newpage

\bibliographystyle{IEEEtran}
\bibliography{IEEEabrv,references}

\end{document}